\documentclass[10pt,twocolumn,letterpaper]{article}
\usepackage{multicol}
\usepackage[pagenumbers]{cvpr} 
\usepackage[table]{xcolor}

\definecolor{cvprblue}{rgb}{0.21,0.49,0.74}
\usepackage[pagebackref,breaklinks,colorlinks,citecolor=cvprblue]{hyperref}
\def\paperID{*****} 
\def\confName{CVPR}
\def\confYear{2024}
\DeclareUnicodeCharacter{0303}{~}

\title{Splatfacto-W: A Nerfstudio Implementation of Gaussian Splatting for Unconstrained Photo Collections.}

\author{Congrong Xu \\
UC Berkeley, ShanghaiTech \\
{\tt\small xucr@berkeley.edu}
\and
Justin Kerr \\
UC Berkeley \\
{\tt\small kerr@berkeley.edu}
\and
Angjoo Kanazawa \\
UC Berkeley \\
{\tt\small kanazawa@eecs.berkeley.edu}
}

\begin{document}
\onecolumn
\maketitle
\begin{figure}[h]
    \centering
    \includegraphics[width=0.4\textwidth]{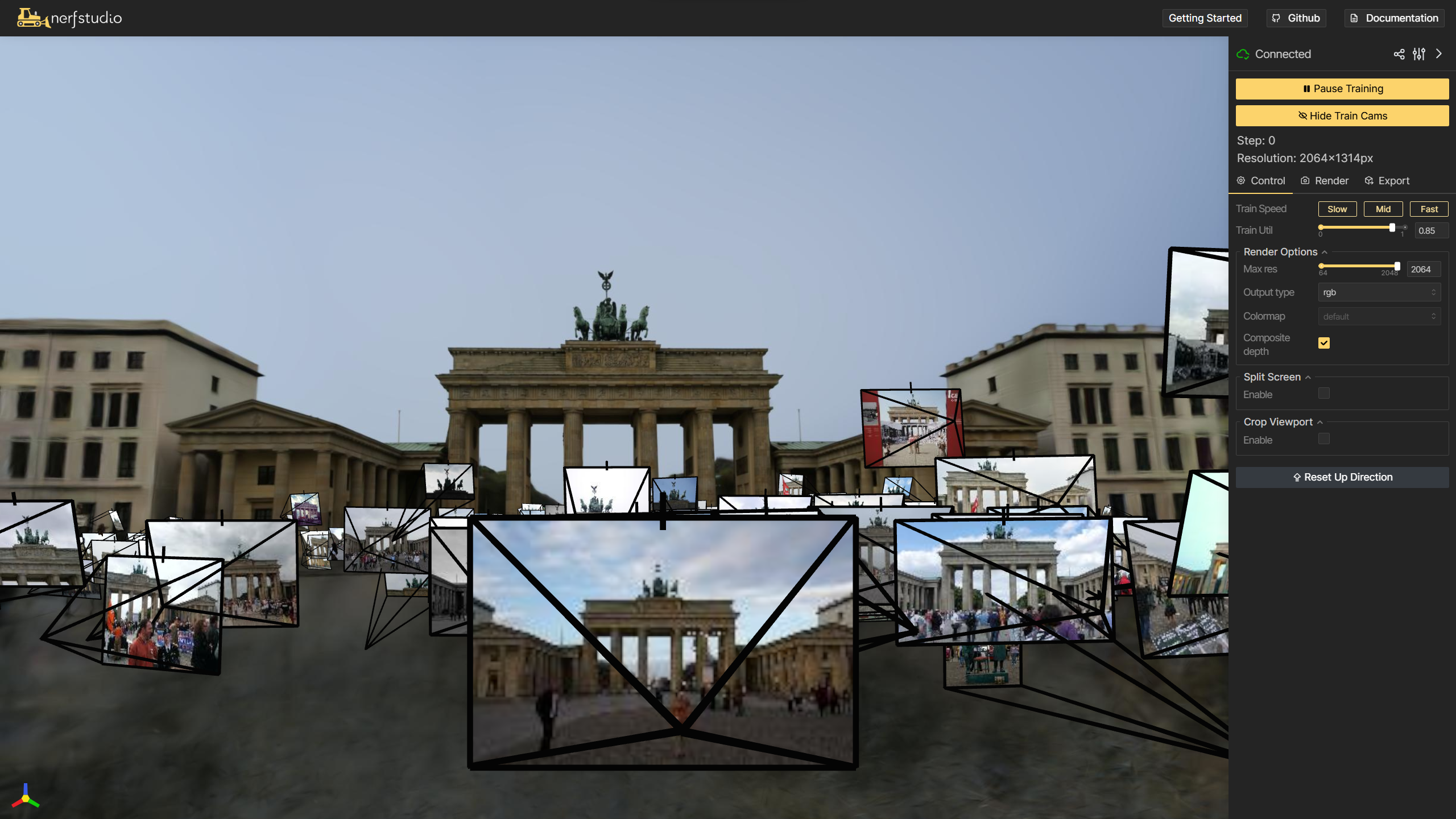}
    \includegraphics[width=0.4\textwidth]{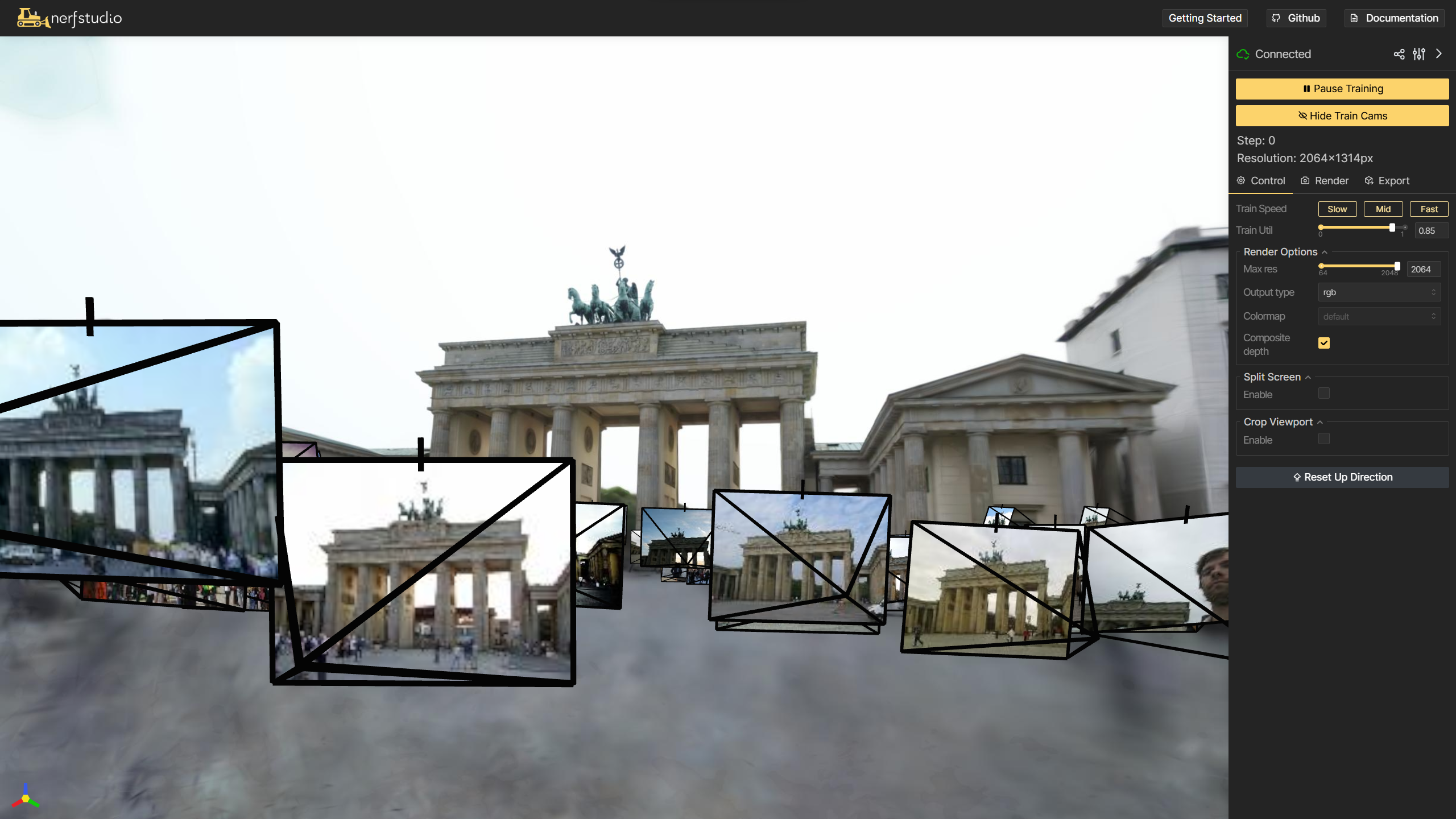}
    \includegraphics[width=0.4\textwidth]{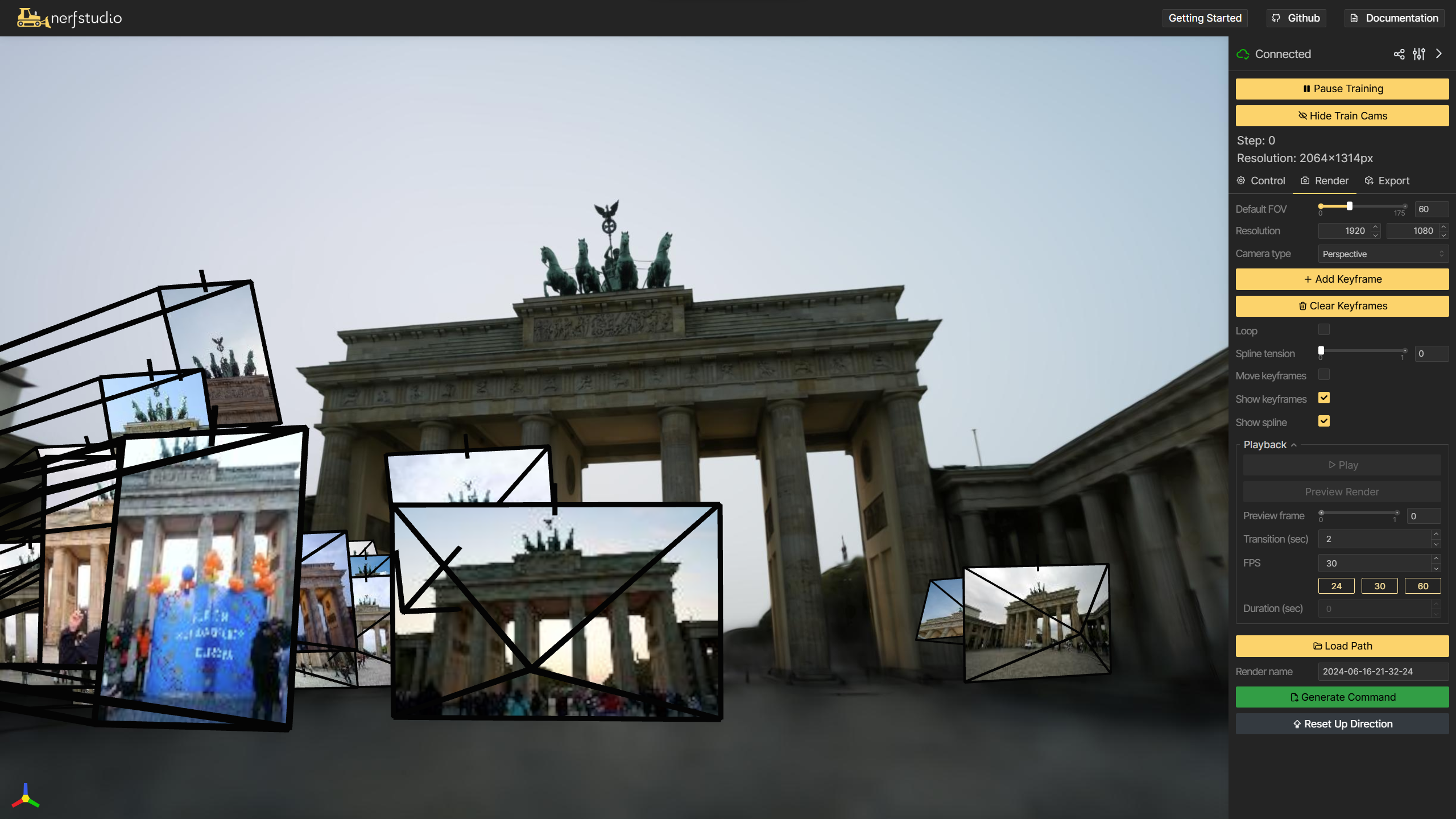}
    \includegraphics[width=0.4\textwidth]{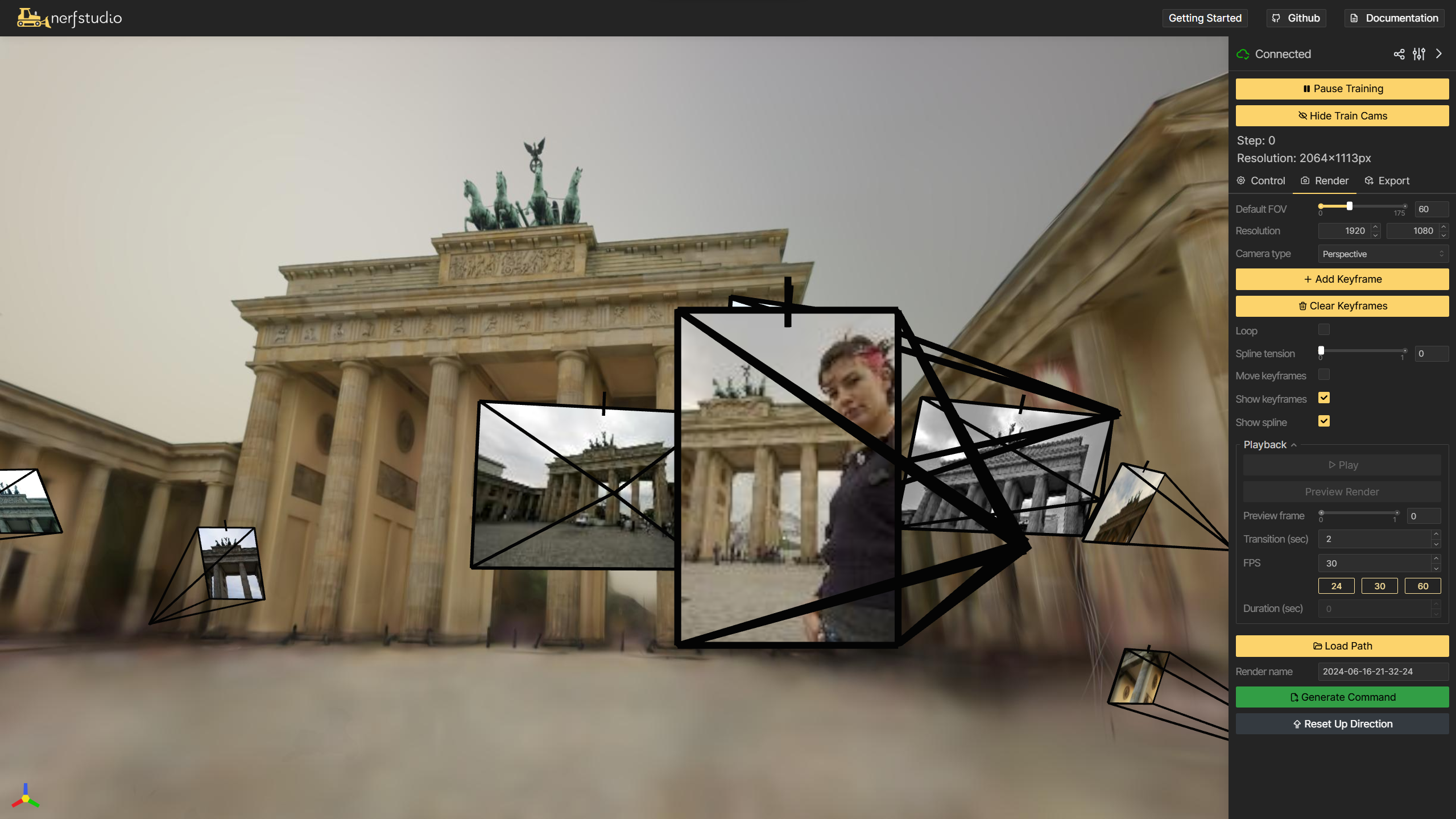}
        
    \caption{\textbf{Splatfacto-W: Real-time exploration of in-the-wild images.} Our approach enables real-time appearance change in the nerfstudio viewer. For example, one can click each input image, and explore the scene using that appearance conditioning, and seamlessly move on to another by clicking an input view. Please see the video on our \href{https://kevinxu02.github.io/splatfactow/}{website}.}
    \label{fig:viewer}
\end{figure}
\begin{multicols}{2}
\begin{abstract}
Novel view synthesis from unconstrained in-the-wild image collections remains a significant yet challenging task due to photometric variations and transient occluders that complicate accurate scene reconstruction. Previous methods have approached these issues by integrating per-image appearance features embeddings in Neural Radiance Fields (NeRFs). Although 3D Gaussian Splatting (3DGS) offers faster training and real-time rendering, adapting it for unconstrained image collections is non-trivial due to the substantially different architecture. 
In this paper, we introduce Splatfacto-W, an approach that integrates per-Gaussian neural color features and per-image appearance embeddings into the rasterization process, along with a spherical harmonics-based background model to represent varying photometric appearances and better depict backgrounds. Our key contributions include latent appearance modeling, efficient transient object handling, and precise background modeling. Splatfacto-W delivers high-quality, real-time novel view synthesis with improved scene consistency in in-the-wild scenarios. Our method improves the Peak Signal-to-Noise Ratio (PSNR) by an average of 5.3 dB compared to 3DGS, enhances training speed by 150 times compared to NeRF-based methods, and achieves a similar rendering speed to 3DGS. Additional video results and code integrated into Nerfstudio are available at \href{https://kevinxu02.github.io/splatfactow/}{https://kevinxu02.github.io/splatfactow/}.
\end{abstract}

\section{Introduction}
Novel view synthesis from a collection of 2D images has garnered significant attention for its wide-ranging applications including virtual reality, augmented reality, and autonomous navigation. Traditional methods such as Structure-from-Motion (SFM) \cite{schonberger2016structure} and Multi-View Stereo (MVS), and more recently Neural Radiance Fields~\cite{mildenhall2020nerf} and its extensions~\cite{barron2021mipnerf,mueller2022instant} have laid the groundwork for 3D scene photometric reconstruction. However, these approaches often struggle with image collections captured at the same location under different appearances, for example time-of-day or weather variations, which exhibit photometric variations, transient occluders, or scene inconsistency. Extensions to NeRF like NeRF-W~\cite{martin2021nerf} or others~\cite{kassab2023refinedfields,kplanes_2023,yang2023cross} are able to capture these variations by optimizing per-image appearance embeddings and conditioning its rendering on these. These methods however are slow to optimize and render,

On the other hand, 3D Gaussian Splatting~\cite{kerbl3Dgaussians} has emerged as a promising alternative, offering faster training and real-time rendering capabilities. 3DGS represents scenes using explicit 3D Gaussian points and employs a differentiable rasterizer to achieve efficient rendering. However, the explicit nature of 3DGS makes handling in-the-wild cases via per-image appearance embedding non-trivial. 

In this paper, we introduce a simple, straightforward approach for handling in-the-wild challenges with 3DGS, called Splatfacto-W, implemented in Nerfstudio. Our method achieves a significant improvement in PSNR, with an average increase of 5.3 dB compared to 3DGS. Splatfacto-W maintains a rendering speed comparable to 3DGS, enabling real-time performance on commodity GPUs such as the RTX 2080Ti. Additionally, our approach effectively handles background representation, addressing a common limitation in 3DGS implementations.

There have been efforts to handle in-the-wild scenarios with 3DGS, such as SWAG \cite{dahmani2024swag} and GS-W \cite{zhang2024gsw}. However, these approaches have limitations. SWAG's implicit color prediction slows down rendering due to the need to query latent embeddings, while GS-W's reliance on 2D models restricts both training and inference speed. In contrast, Splatfacto-W offers several key contributions:

\begin{enumerate}
    \item \textbf{Latent Appearance Modeling}: We assign appearance feature for each Gaussian point, enabling effective Gaussian color adaptation to variations in reference images. This can later be converted to explicit colors, ensuring the rendering speed.
    \item \textbf{Transient Object Handling}: An efficient heuristic based method for masking transient objects during the optimization process, improving the focus on consistent scene features, without reliance on 2D pretrained models.
    \item \textbf{Background Modeling}: A spherical harmonics-based background model that accurately represents the sky and background elements, ensuring improved multi-view consistency.
\end{enumerate}
Our approach can handle in-the-wild challenges such as diverse lighting at PSNR 17\% higher than NeRF-W, while enabling Real-Time interaction, as illustrated in Figure \ref{fig:viewer}, and videos on our \href{https://kevinxu02.github.io/splatfactow/}{website}.

\section{Related Work}
\subsection{Neural Rendering in the Wild}
Pioneering approaches such as NeRF-W~\cite{martin2021nerf}, proposed disentangling static and transient occluders by employing two per-image embeddings (appearance and transient) alongside separate radiance fields for the static and transient components of the scene. In contrast, Ha-NeRF~\cite{chen2022hallucinated} uses a 2D image-dependent visibility map to eliminate occluders, bypassing the need for a decoupled radiance field since transient phenomena are only observed in individual 2D images. This simplification helps reduce the blurry artifacts encountered by NeRF-W ~\cite{martin2021nerf} when reconstructing transient phenomena with a 3D transient field.

Building on previous methods, CR-NeRF~\cite{yang2023cross} improves performance by leveraging interaction information from multiple rays and integrating it into global information. This method employs a lightweight segmentation network to learn a visibility map without the need for ground truth segmentation masks, effectively eliminating transient parts in 2D images. Another recent advancement, RefinedFields~\cite{kassab2023refinedfields}, utilizes K-Planes and generative priors for in-the-wild scenarios. This approach alternates between two stages: scene fitting to optimize the K-Planes~\cite{kplanes_2023} representation and scene enrichment to finetune a pre-trained generative prior and infer a new K-Planes representation.

Implicit-field representations have seen diverse adaptations for in-the-wild scenarios. However, their training and inference processes are time-consuming, posing a significant challenge to achieving real-time rendering. This limitation hinders their application in practical scenarios where fast rendering speed is essential, particularly in various interactive 3D applications. Inspired by the appearance embeddings in NeRF-W, Splatfacto-W uses a image wise appearance embedding to handle lighting variations.
\subsection{Gaussian Splatting in the Wild}
Recent advancements in 3D Gaussian Splatting (3DGS)~\cite{kerbl3Dgaussians} have shown promise for efficient and high-quality novel view synthesis, particularly for static scenes. However, the challenge remains to adapt these methods for unconstrained, in-the-wild image collections that include photometric variations and transient occluders. Two significant contributions to this field are SWAG (Splatting in the Wild images with Appearance-conditioned Gaussians)~\cite{dahmani2024swag} and GS-W (Gaussian in the Wild)~\cite{zhang2024gsw}.

SWAG~\cite{dahmani2024swag} extends 3DGS by introducing appearance-conditioned Gaussians. This method models the appearance variations in the rendered images by learning per-image embeddings that modulate the colors of the Gaussians via a multilayer perceptron (MLP). Additionally, SWAG addresses transient occluders using a new mechanism that trains transient Gaussians in an unsupervised manner, improving the scene reconstruction quality and rendering efficiency compared to previous methods~\cite{martin2021nerf,kplanes_2023,chen2022hallucinated}. However, the color prediction for each Gaussian in SWAG is implicit, requiring a query of the latent embedding for each Gaussian in the hash grid, which slows down the rendering speed to about 15 FPS form the 181 FPS of 3DGS.

Similarly, GS-W~\cite{zhang2024gsw} proposes enhancements for handling in-the-wild scenarios by equipping each 3D Gaussian point with separate intrinsic and dynamic appearance features. This separation allows GS-W to better model the unique material attributes and environmental impacts for each point in the scene. Moreover, GS-W introduces an adaptive sampling strategy to capture local and detailed information more effectively and employs a 2D visibility map to mitigate the impact of transient occluders. However, this method introduces 2D U-Nets that slow down both the training and inference speed, and it also limits the rendering resolution.

Our method improves on the speed limitations of both SWAG and GS-W and introduces a spherical harmonics based background model to address the background issue, ensuring improved multiview consistency.
\begin{figure*}[t]
\centering
  \includegraphics[width=\textwidth]{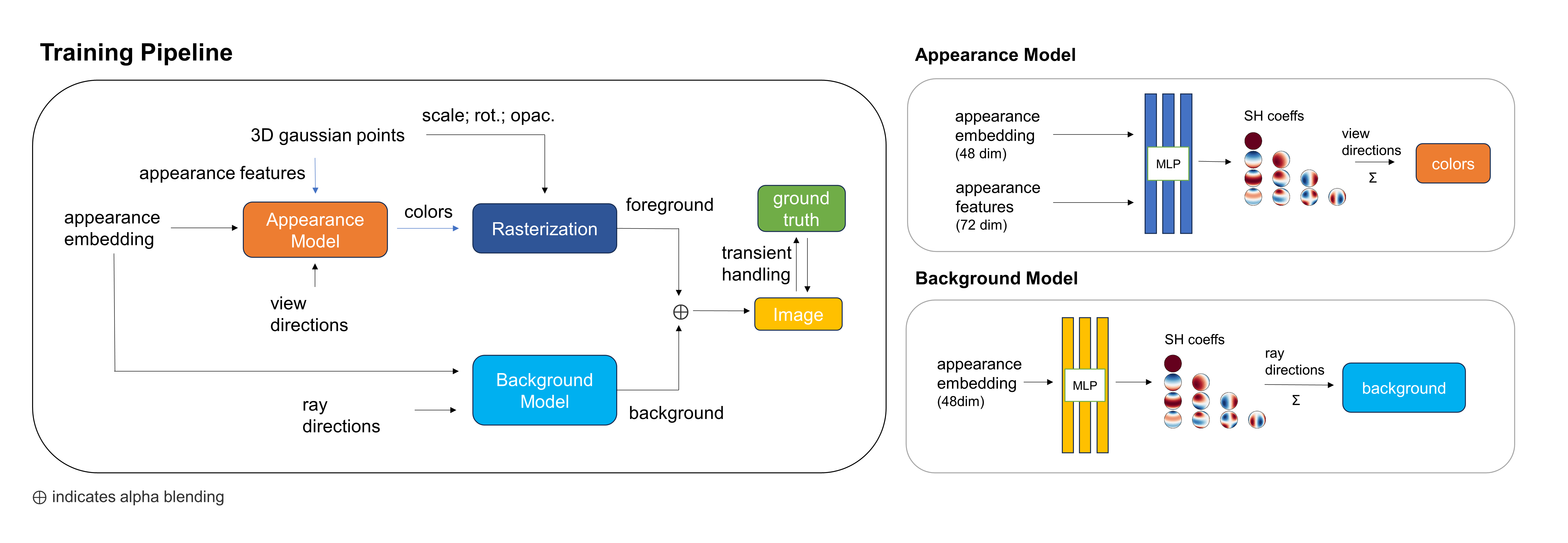}
  \caption{We begin by predicting the color of each Gaussian using the Appearance Model. These Gaussians are then rasterized to generate the foreground objects. While Background Model predicts the background given ray directions. The foreground and background are merged using alpha blending to produce the final image. This final image is compared with the masked ground truth image and then processed through the Robust Mask to update the model parameters.}
  \label{fig:Pipeline}
\end{figure*}

\section{Preliminaries}
3D Gaussian Splatting \cite{kerbl3Dgaussians} is a method for reconstructing 3D scenes from static images with known camera poses. It represents the scene using explicit 3D Gaussian points (Gaussians) and achieves real-time image rendering through a differentiable tile-based rasterizer. The positions ($\mu$) of these Gaussian points are initialized with point clouds extracted by Structure-from-Motion (SFM) \cite{schonberger2016structure} from the image set. 

The 3D covariance ($\Sigma$) models the influence of each Gaussian point on the color anisotropy of the surrounding area:
\begin{equation}
G(x - \mu, \Sigma) = e^{-\frac{1}{2} (x - \mu)^T \Sigma^{-1} (x - \mu)}
\end{equation}


Each Gaussian point is also equipped with opacity ($\alpha$) and color ($c$) attributes, with the color represented by third-order spherical harmonic coefficients. When rendering, the 3D covariance ($\Sigma$) is projected to 2D ($\Sigma'$) using the viewing transformation ($W$) and the Jacobian of the affine approximation of the projective transformation ($J$):
\begin{equation}
\Sigma' = JW \Sigma W^T J^T
\end{equation}

The color of each pixel is aggregated using $\alpha$-blending:
\begin{equation}
\sigma_i = G(px' - \mu_i, \Sigma'_i)
\end{equation}
\begin{equation}
C(r) = \sum_{i \in G_{r}} c_i \sigma_i \prod_{j=1}^{i-1} (1 - \sigma_j)
\end{equation}

Here, $r$ represents the position of a pixel, and $G_{r}$ denotes the sorted Gaussian points associated with that pixel. The final rendered image is used to compute the loss with reference images for training, optimizing all Gaussian attributes. Additionally, a strategy for point growth and pruning based on gradients and opacity is devised.

\section{Splatfacto-W}
\label{sec:headings}
We now present Splatfacto-W, a system for reconstructing 3D scenes from in-the-wild photo collections. We build on top of Splatfacto in Nerfstudio~\cite{nerfstudio} and introduce three modules explicitly designed to handle the challenges of unconstrained imagery. An illustration of the whole pipeline can be found in Fig.~\ref{fig:Pipeline}.

\subsection{Latent Appearance Modeling}
3D Gaussian Splatting \cite{kerbl3Dgaussians} is designed for reconstructing scenes from consistent image sets and employs spherical harmonic coefficients for color modeling. In our approach, we deviate from this convention. Instead, we introduce a new appearance feature $f_i$ for each Gaussian point, adapting to variations in the reference images along with the appearance embedding vector $\ell_j$ of dimension $n$.

We predict the spherical harmonics coefficients $\mathbf{b}_i$ of each Gaussian using a multi-layer perceptron (MLP), parameterized by $\theta$:
\[
\mathbf{b}_i = \text{MLP}_{\theta}(\ell_j, f_i)
\]
where $\mathbf{b}_i = (b_{i,\ell}^m) {0 \leq \ell \leq \ell{\text{max}}, -\ell \leq m \leq \ell}$.

The $\{\ell_j\}_{j=1}^{N_{img}}$ and $\{fi\}_{i=1}^{N{gs}}$
embeddings are optimized alongside $\theta$, where $N_{img}$ is the number of images and ${N_{gs}}$ is the number of gaussian points. 

We then recover the color $c_i$ for gaussian point $i$ from the SH coeffcients $\mathbf{b}_i$:
\[
c_i = \text{Sigmoid}\left(\sum_{\ell=0}^{\ell_{\text{max}}} \sum_{m=-\ell}^{\ell} b_{i,\ell}^m Y_\ell^m(\mathbf{d}_i)\right)
\]
Here, $\mathbf{d}_i$ is the viewing direction for gaussian point $i$. $Y_\ell^m$ are the spherical harmonic basis functions. 

This approach allows us to prevent inputting the viewing directions into the MLP, hence we can cache the gaussian status with a single inference for any appearance embedding, allowing us to have the same rendering speed as 3DGS. 


\subsection{Transient Handling with Robust Mask}
Our objective is to develop an efficient method for mask creation that addresses transient objects within the optimization process of Gaussian Splatting. Gaussian Splatting's dependence on initialized point clouds results in suboptimal performance for transient object representation, leading to increased loss in affected regions. By strategically masking pixels, we aim to enhance the model's focus on more consistent scene features.

We adopt a strategy similar to RobustNeRF \cite{sabour2023robustnerf}. We hypothesize that residuals surpassing a certain percentile between the ground truth and the rendered image indicate transient objects, and thus, their corresponding pixels should be masked.

Additionally, we posit that a lower loss between the ground truth and the predicted image signifies a more accurate representation, implying fewer transient objects.

According to the previous assumption, we record the maximum, minimum, and the current $L1$ loss between the ground truth image and the predicted image before any masking. We then linearly interpolate the current mask percentage between the maximum and minimum masking percentages ($Per_{max}$ and $Per_{min}$).

As optimization progresses, images with fewer transient objects exhibit lower loss, thereby reducing the mask percentage. Conversely, images with more transient objects retain higher loss.

The threshold for masking is determined as follows.
\[
\mathcal{T}_{\epsilon} = (1 - k)\% \text{ percentile of residuals for all pixels}
\]
where
\[
k = \frac{L1_{current} - L1_{min}}{L1_{max} - L1_{min}} \times (Per_{max} - Per_{min}) + Per_{min}
\]

We start by creating a per-pixel mask $\tilde{\omega}(\mathbf{r})$, where inlier (i.e., pixels to be learned by the model) is 1 and outlier (i.e., pixels to be masked and not learned by the model) is 0.

To ensure more efficient model convergence, we introduce an additional condition: always mark the pixels belonging to the upper n\% of the image as inlier, as this region typically corresponds to the sky in most images. We define an upper n\% (choosing n=40 in practice) region mask:

\[
U(\mathbf{r}) = \begin{cases}
1 & \text{if } r_y \leq 0.4H, \\
0 & \text{otherwise},
\end{cases}
\]
where \( H \) is the height of the image and \( r_y \) is the row coordinate of a pixel.

Thus, \( \tilde{\omega}(\mathbf{r}) \) is activated (marking the pixel as inlier) when the loss \( \epsilon(\mathbf{r}) \) at pixel \( \mathbf{r} \) is less than or equal to \( \mathcal{T}_{\epsilon} \) or belongs to the upper 40\% of the image.
\[
\tilde{\omega}(\mathbf{r}) = (\epsilon(\mathbf{r}) \leq \mathcal{T}_{\epsilon}) \lor U(\mathbf{r}),
\]
where \(\lor\) denotes the logical OR operation.

Furthermore, to capture the spatial smoothness of transient objects, we spatially blur inlier/outlier labels $\tilde{\omega}$ with a $5 \times 5$ box kernel $\mathcal{B}_{5 \times 5}$. The final mask $\mathcal{W}$ is expressed as:
\[
\mathcal{W}(\mathbf{r}) = (\tilde{\omega} \ast \mathcal{B}_{5 \times 5})(\mathbf{r}) \geq \mathcal{T}_{\ast}, \quad \mathcal{T}_{\ast} = 0.4.
\]
This tends to remove high-frequency details from being classified as pixels for transient objects, allowing them to be captured during optimization.

\subsection{Background Modeling}
Since 3DGS lacks depth perception for images and outdoor images often feature large areas of solid color in the background, it is challenging to accurately represent the background in outdoor scenes. Furthermore, the initial point cloud inadequately represents the spatial positions of the sky. This leads to inconsistent representation of the sky during the 3DGS optimization process, where sky elements may appear close to the camera or adjacent to building structures and tree leaves. This occurs as new Gaussians, intended to represent the background, are split from those representing foreground objects, resulting in a scattered and inaccurate depiction of the sky and overall background.

Moreover, images from in-the-wild collections exhibit varied appearances of the sky, further exacerbating this issue. Since 3DGS focuses only on image space matching, the sky often connects with the optimized scene structure, thereby losing multi-view consistency.

Although we can introduce 2D depth model priors or background segmentation to force the Gaussians to represent the background in the distance, this undoubtedly increases the computational overhead and additional model dependency. Furthermore, it is unwise to use tens of thousands of Gaussians to represent relatively simple background parts of the image.

To address this issue, we introduce a simple yet effective prior: the background should be represented at infinity. Given that the sky portion is typically characterized by low-frequency variations, we found that using only three levels of Spherical Harmonics (SH) basis functions can accurately model the sky.
For scenes with consistent backgrounds, we can directly optimize a set of SH coefficients $\mathbf{b}$ to efficiently model the background. 

However, in in-the-wild scenarios, backgrounds often vary across different images. To accommodate this variability, we employ a Multi-Layer Perceptron (MLP) that takes an appearance embedding vector $\ell_j$ as input and predicts the SH coefficients $\mathbf{b}$ for the background of the current image:
\[
\mathbf{b} = \text{MLP}(\ell_j),
\]
where $\mathbf{b} = (b_\ell^m){0 \leq \ell \leq \ell{\text{max}}, -\ell \leq m \leq \ell}$.

We then derive the color of the sky at infinity for each pixel's ray direction $\mathbf{d}_{\text{ray}}(\mathbf{r})$. For a pixel at position $\mathbf{r}$, the background color $C_{\text{background}}(\mathbf{r})$ is predicted as:
\[
C_{\text{background}}(\mathbf{r}) = \text{Sigmoid}\left(\sum_{\ell=0}^{\ell_{\text{max}}} \sum_{m=-\ell}^{\ell} b_\ell^m Y_\ell^m(\mathbf{d}_{\text{ray}}(\mathbf{r}))\right),
\]
where $Y_\ell^m$ are the spherical harmonic basis functions.

To compute the final color for each pixel, we use alpha blending between the foreground color $C(\mathbf{r})$ and the background color:
\[
C_{\text{final}}(\mathbf{r}) = C(\mathbf{r}) + (1 - \alpha(\mathbf{r})) C_{\text{background}}(\mathbf{r})
\]
where $\alpha(\mathbf{r})$ is the alpha value (opacity) at pixel position $\mathbf{r}$.

Furthermore, we introduce a new loss term: the alpha loss. This loss is designed to penalize Gaussians (representing potential foreground objects) that incorrectly occupy pixels well represented by the background model. 

We start by picking out pixels \(p_i\) are well presented by the background model (i.e., the residual between the background and the ground truth is below a certain threshold). To avoid false positives and utilize the low frequency nature of the background, we ensure that the surrounding pixels of each selected pixel also belong to the background. Otherwise, we deselect that pixel.

We encourage the alpha of the gaussians corresponding to these pixels to be low. Specifically, the alpha loss \( L_{\alpha} \) can be expressed as:
\[
    L_{\alpha} = \lambda \times \sum_{\mathbf{r} \in p_i} \alpha(\mathbf{r})
\]
where \(\alpha(\mathbf{r})\) is the accumulation of the Gaussians at pixel \(r\), and \(\lambda\) is a scaling factor. The set \(p_i\) is defined as:
\[
p_i = \{ (\mathbf{r}') : M'(\mathbf{r}') > 0.6 \}
\]
where \(M'(\mathbf{r})\) is the result of applying a \(3 \times 3\) box filter to the residual mask \(M\), computed as:
\[
M(r) = 1_{|\text{Ground Truth}(\mathbf{r}) - \text{Predicted Background}(\mathbf{r})| < \text{Threshold}}
\]
and
\[
M'(\mathbf{r}) = (M \ast \mathcal{B}_{3 \times 3})(\mathbf{r})
\]
This approach considers the smoothness of the background, ensures that only those pixels significantly represented by the background model, as confirmed by the filtered mask, contribute to the alpha loss.

    
        

\begin{figure*}[t]
    \centering
        \includegraphics[width=0.4\textwidth]{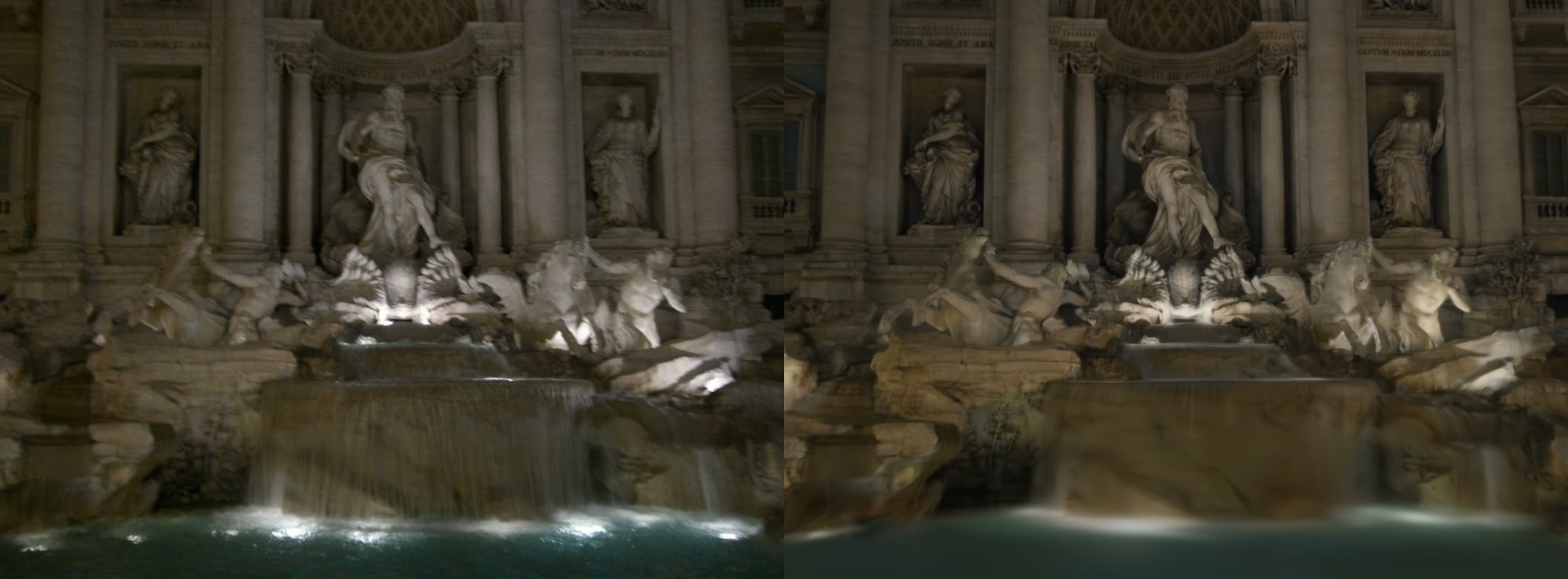}
        \includegraphics[width=0.4\textwidth]{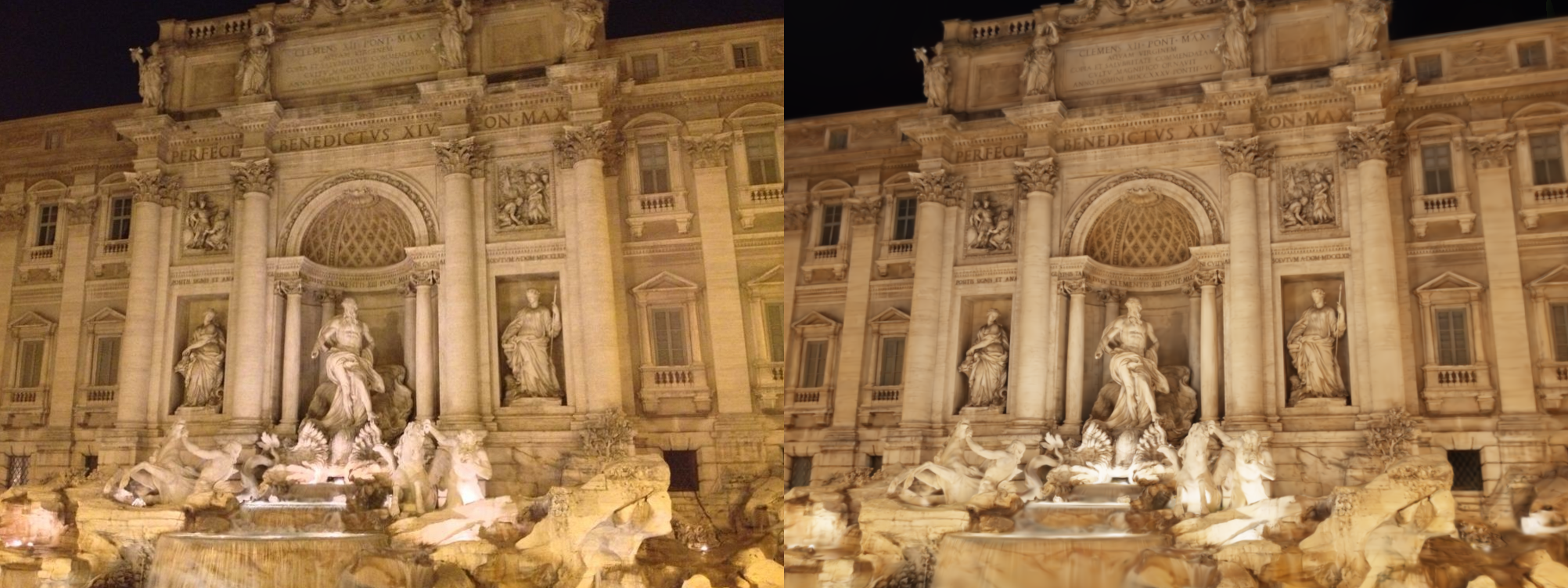}
        \centering
        \includegraphics[width=0.4\textwidth]{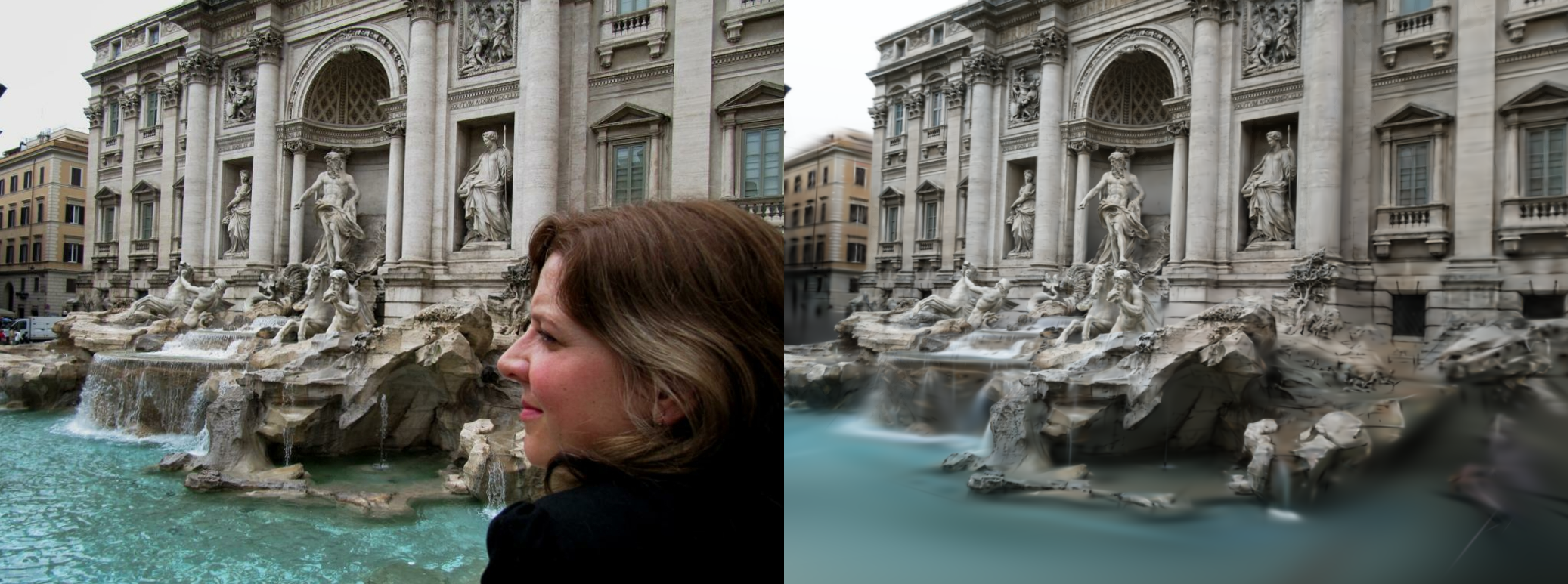}
        \includegraphics[width=0.4\textwidth]{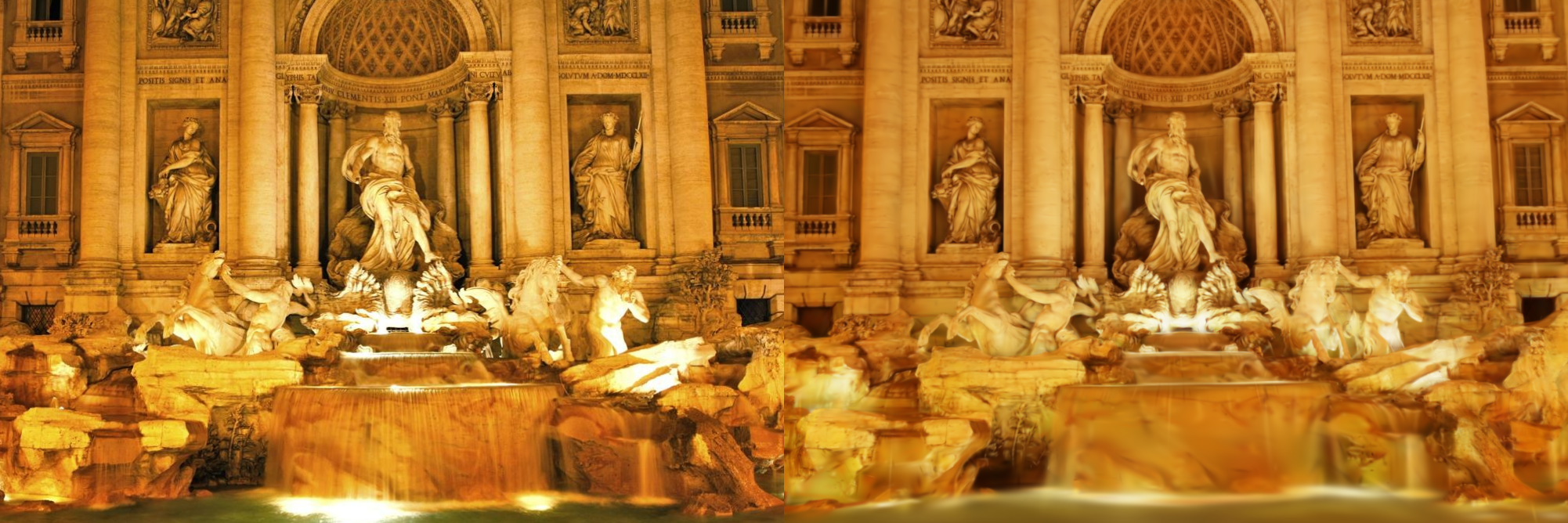}
        \centering
        \includegraphics[width=0.4\textwidth]{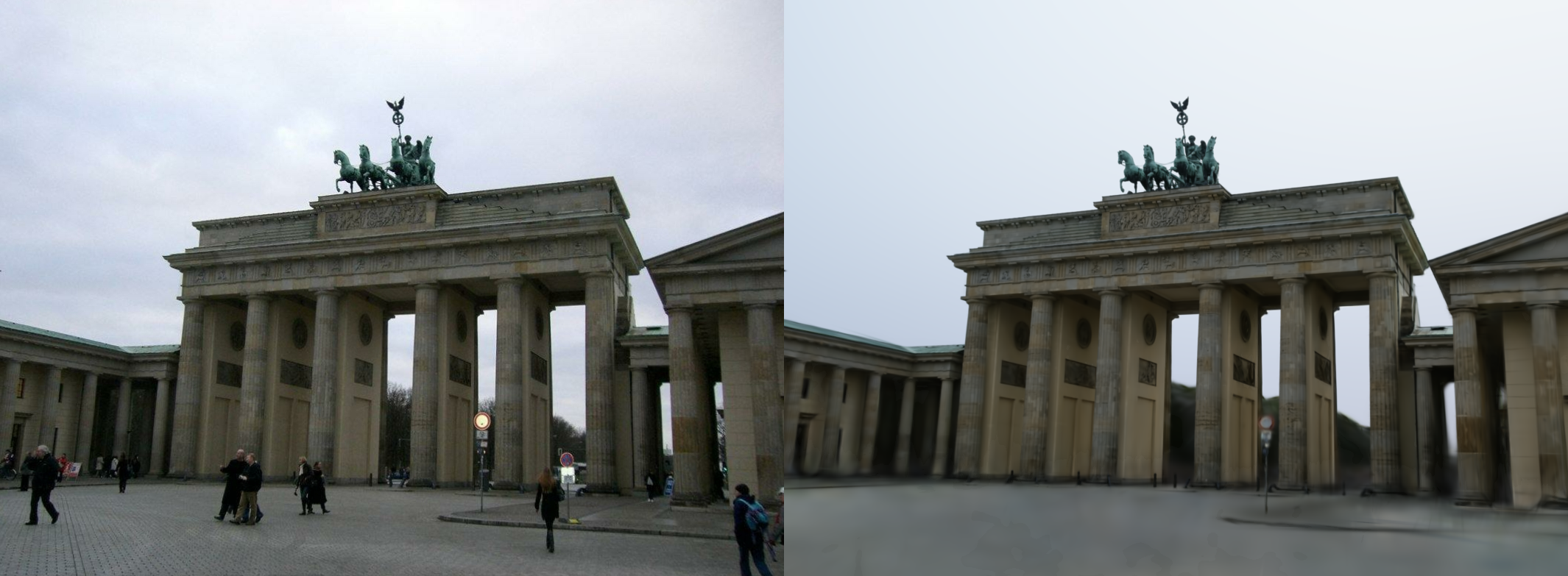}
        \includegraphics[width=0.4\textwidth]{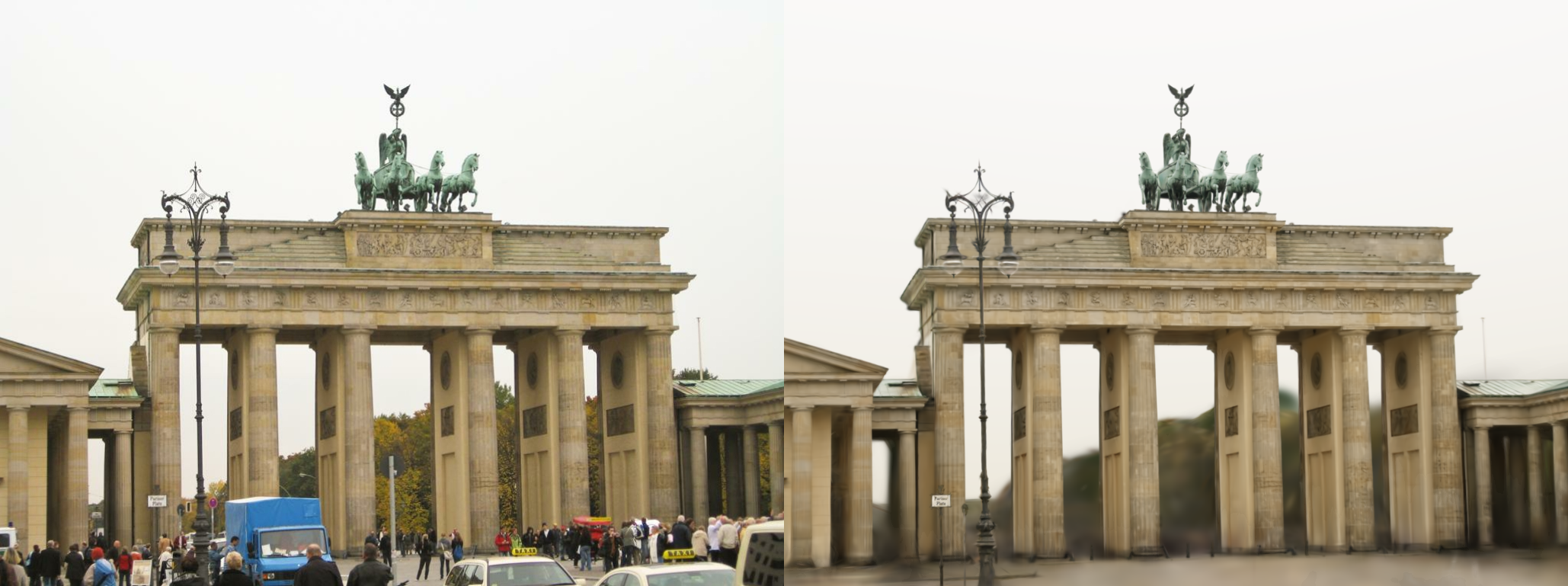}
        \centering
        \includegraphics[width=0.4\textwidth]{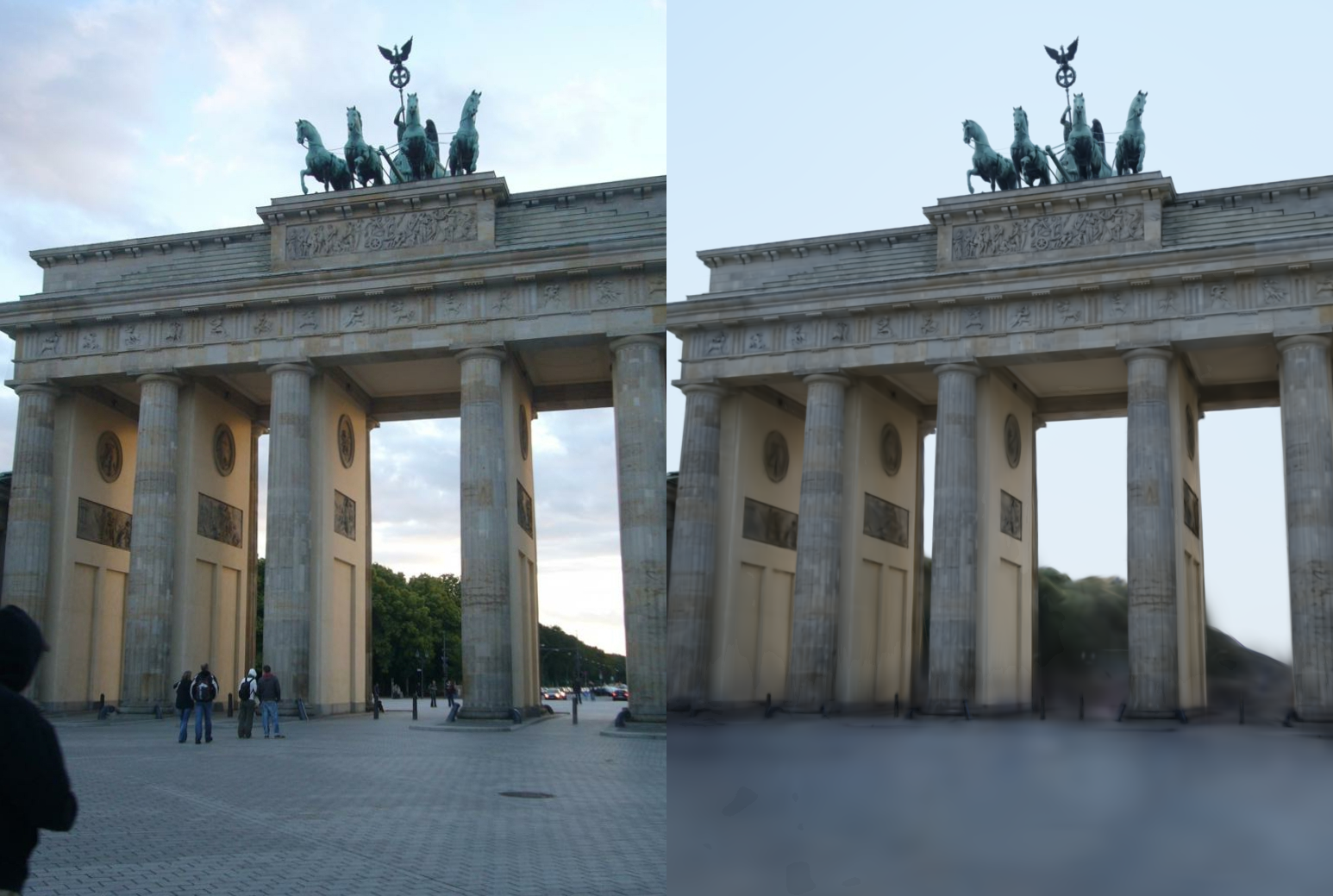}
        \includegraphics[width=0.4\textwidth]{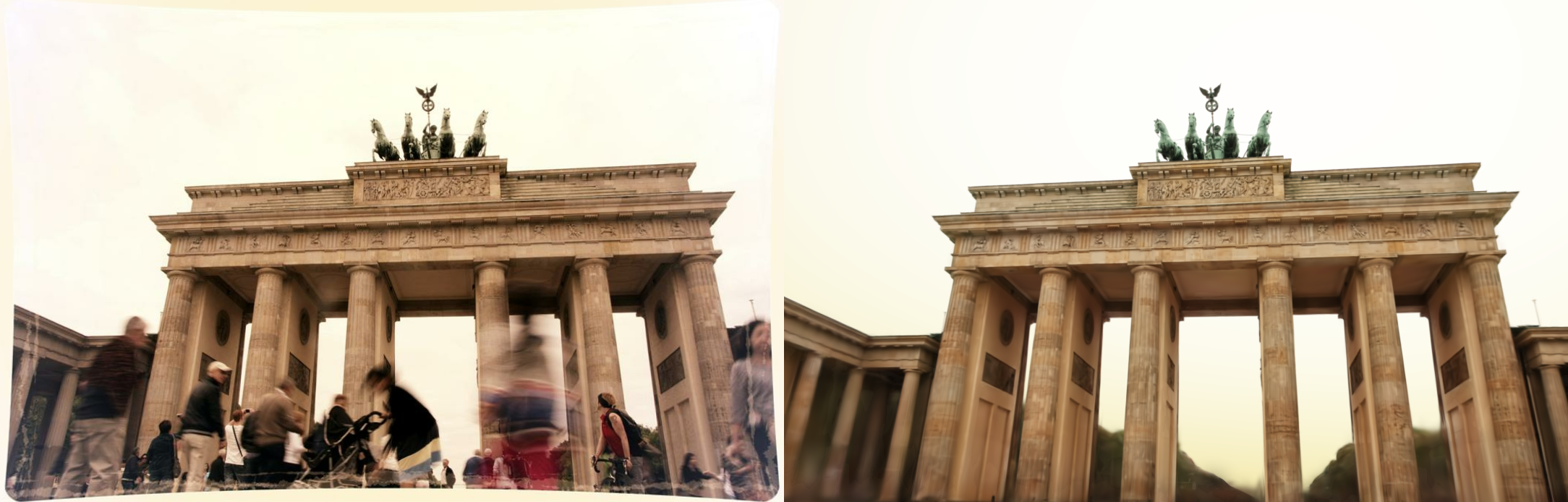}
        \centering
        \includegraphics[width=0.4\textwidth]{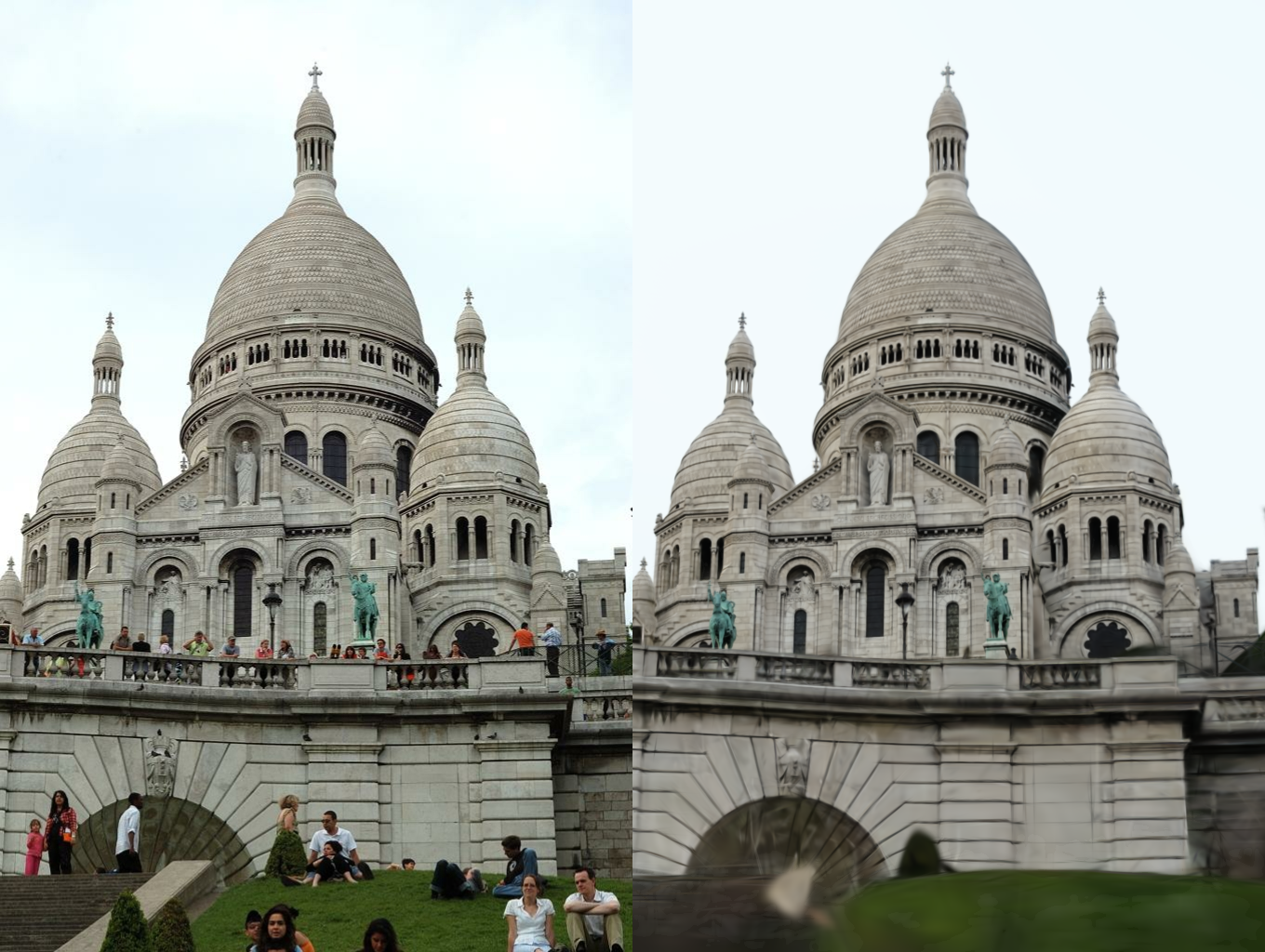}
        \includegraphics[width=0.4\textwidth]{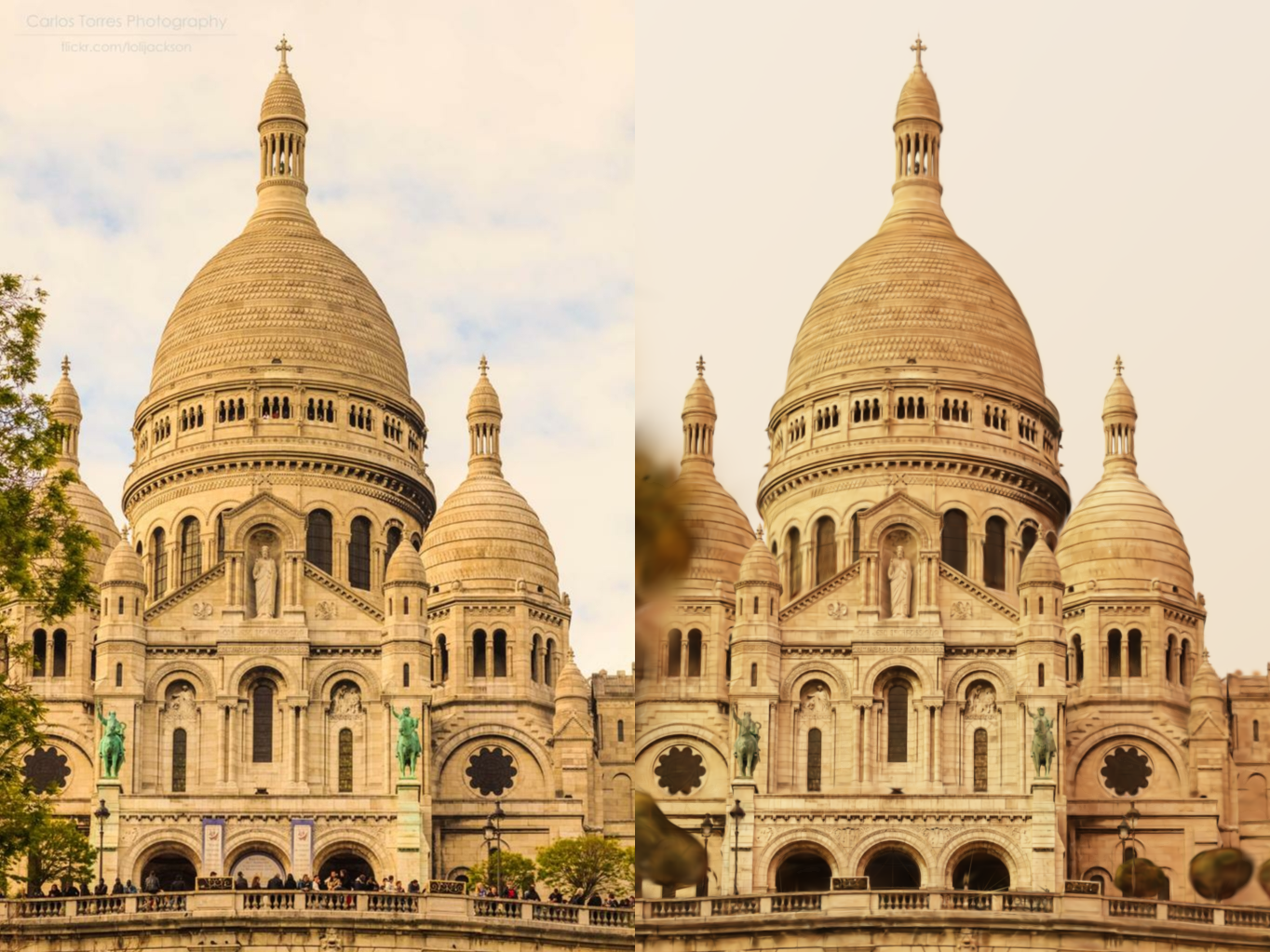}
        \centering
        \includegraphics[width=0.4\textwidth]{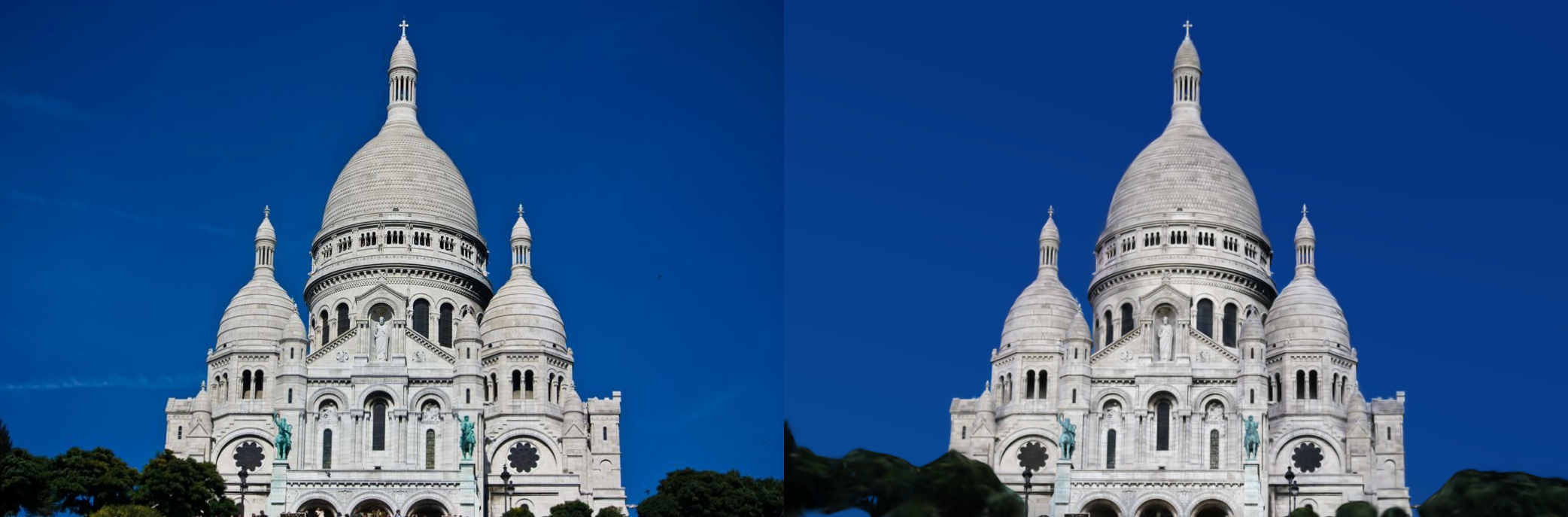}
        \includegraphics[width=0.4\textwidth]{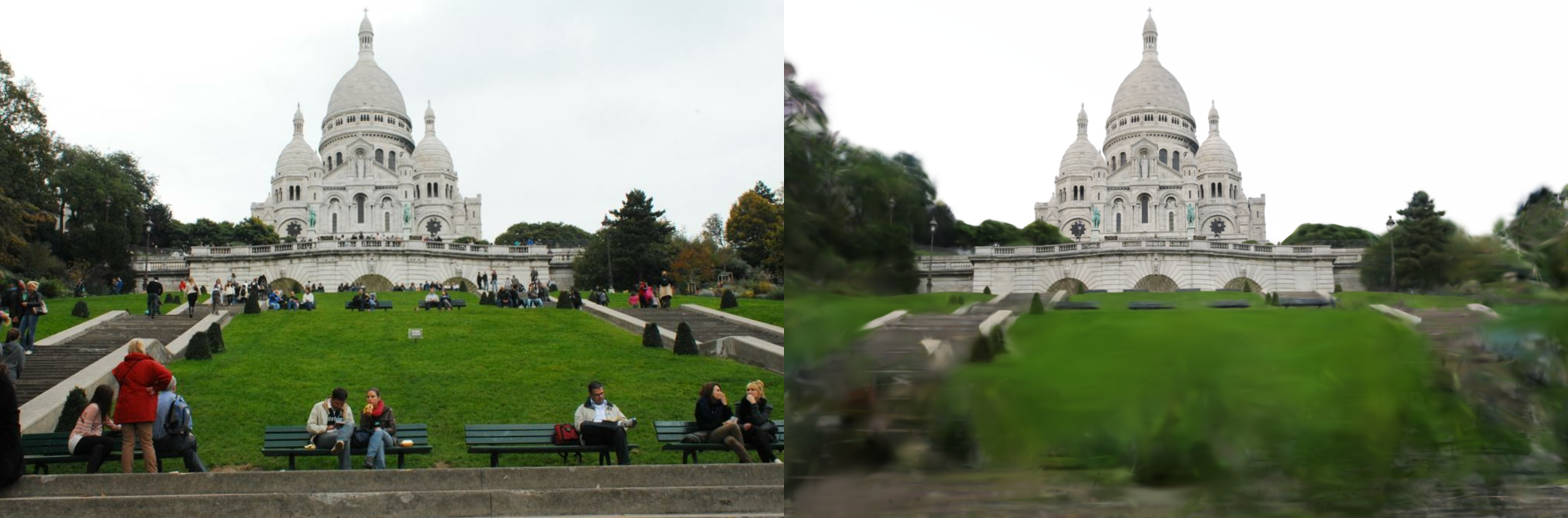}
    \caption{\textbf{Eval Results for Trevi Fountain, Brandenburg Gate, and Sacre Coeur} (Left: Ground Truth; Right: Splatfacto-W)}
    \label{fig:scenes}
    \hfill
\end{figure*}
\newpage
\begin{figure*}[t]
    \centering
    \includegraphics[width=0.8\textwidth]{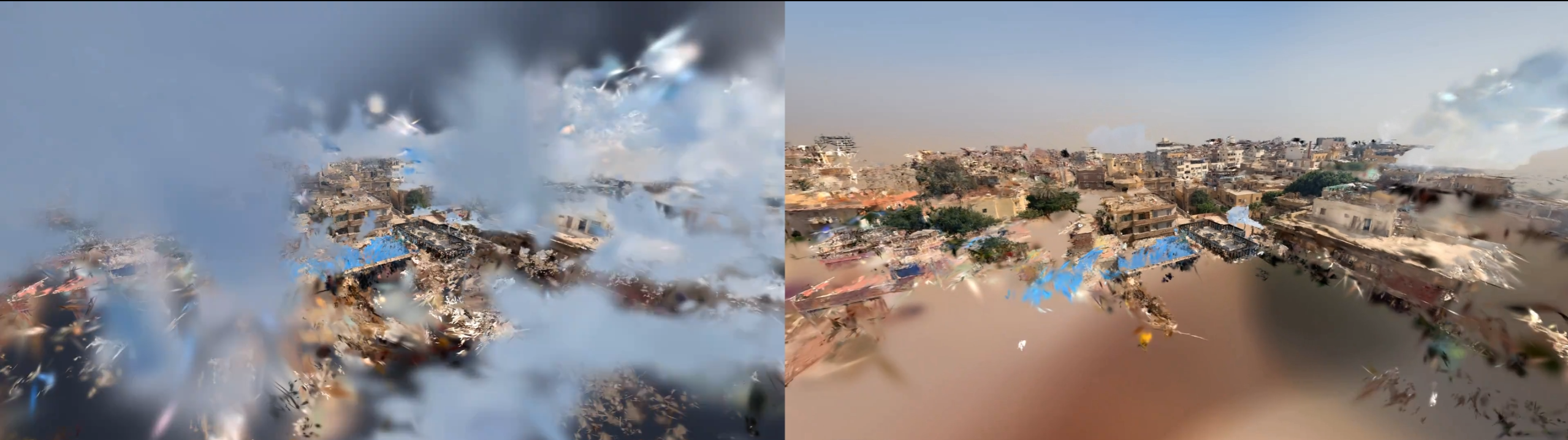}
    \includegraphics[width=0.8\textwidth]{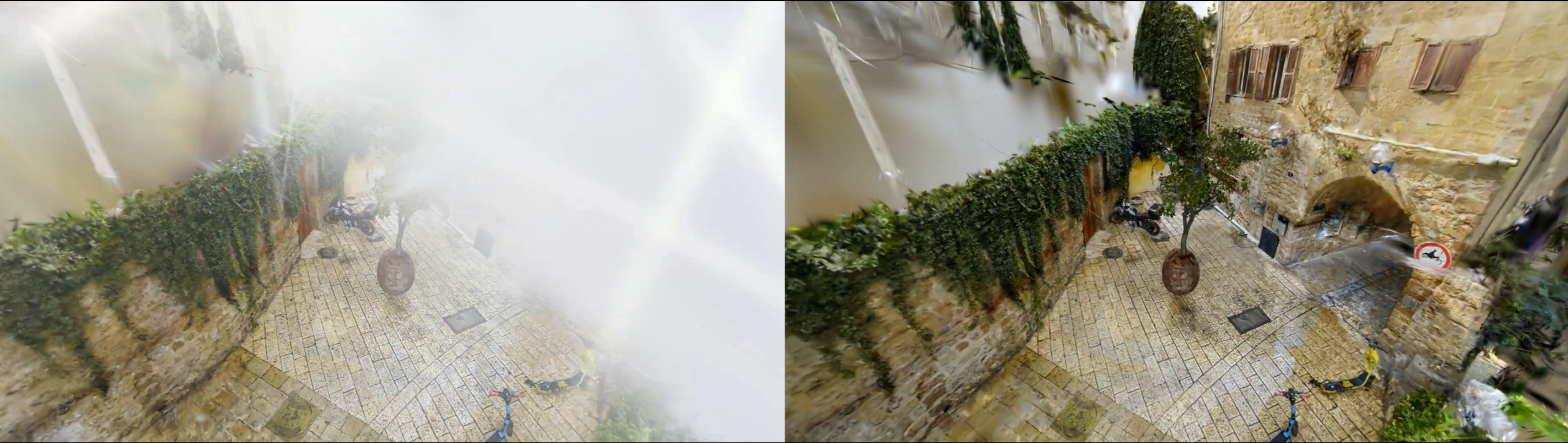}
    \includegraphics[width=0.8\textwidth]{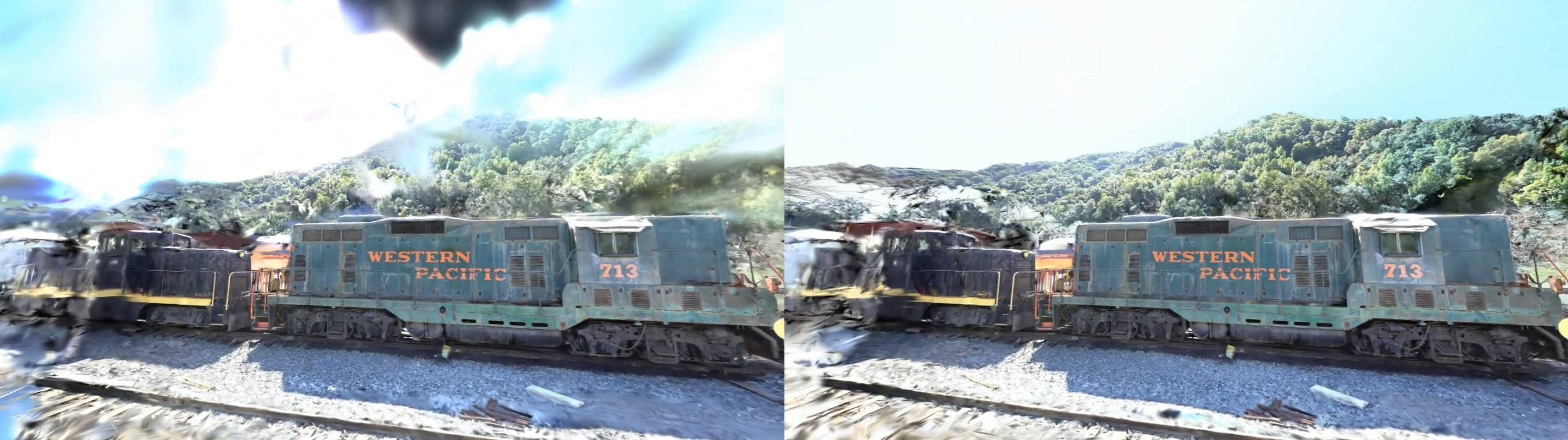}
    \includegraphics[width=0.8\textwidth]{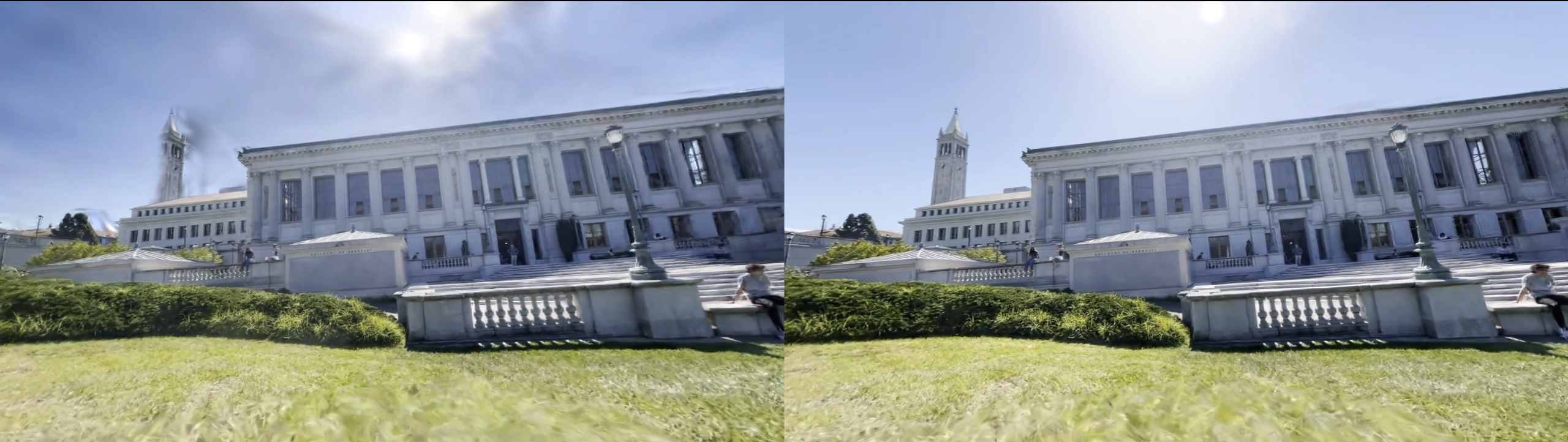}
    \caption{\textbf{Background Modeling in Splatfacto} (Left: Without Background Model; Right: With Background Model)}
    \label{fig:bg}
\end{figure*}

\section{Experiments}
\definecolor{bestcolor}{RGB}{255,150,150}
\definecolor{secondbestc}{RGB}{255,200,130}
\definecolor{thirdbestc}{RGB}{255,255,130}
\begin{table*}[t]
\centering
\resizebox{2\columnwidth}{!}{
\begin{tabular}{lcccccccccccc}
\toprule
& \multicolumn{3}{c}{Brandenburg Gate} & \multicolumn{3}{c}{Trevi Fountain} & \multicolumn{3}{c}{Sacre Coeur} & \multicolumn{2}{c}{Mean Efficiency} \\
\cmidrule(lr){2-4} \cmidrule(lr){5-7} \cmidrule(lr){8-10}\cmidrule(lr){11-12} 
& PSNR ↑ & SSIM ↑ & LPIPS ↓ & PSNR ↑ & SSIM ↑ & LPIPS ↓ & PSNR ↑ & SSIM ↑ & LPIPS ↓& Training Time (h) & FPS \\
\midrule
NeRF~\cite{mildenhall2020nerf}  & 18.90 & 0.815 & 0.231 & 15.60 & 0.715 & 0.291 & 16.14 & 0.600 & 0.366 & - & - \\
NeRF-W~\cite{martin2021nerf}  & 24.17 & 0.890 & 0.167 & 18.97 & 0.698 & 0.265 & 19.20 & 0.807 & 0.191  & 400 & $<$1\\
Ha-NeRF~\cite{chen2022hallucinated}  & 24.04 & 0.877 & 0.139 & 20.18 & 0.690 & 0.222 & 20.02 & 0.801 & 0.171 & 452 & 0.20 \\
CR-NeRF~\cite{yang2023cross} & 26.53 & 0.900 & \cellcolor{secondbestc}0.106 & 21.48 & 0.711 & \cellcolor{secondbestc}0.206 & 22.07 & 0.823 & \cellcolor{secondbestc}0.152 & 420 & 0.25\\
RefinedFields~\cite{kassab2023refinedfields}  & 26.64 & 0.886 & - & \cellcolor{bestcolor}23.42 & 0.737 & - & 22.26 & 0.817 & - & 150 & $<$1 \\
3DGS~\cite{kerbl3Dgaussians} & 19.99 & 0.889 & 0.180 & 18.47 &0.761 & 0.234& 17.57 & 0.831 & 0.219  & 0.30* & 181* \\
SWAG~\cite{dahmani2024swag} & 26.33 & \cellcolor{thirdbestc} 0.929 & 0.139 & \cellcolor{secondbestc}23.10 & \cellcolor{bestcolor}0.815 & \cellcolor{thirdbestc}0.208 & 21.16 & 0.860 & 0.185 &0.83*&15.29*\\
GS-W \cite{zhang2024gsw} & \cellcolor{bestcolor}27.96 & \cellcolor{bestcolor}0.932 & \cellcolor{bestcolor}0.086 & \cellcolor{thirdbestc}22.91 & \cellcolor{secondbestc}0.801 & \cellcolor{bestcolor}0.156 &  \cellcolor{bestcolor}23.24 & \cellcolor{thirdbestc}0.863 & \cellcolor{bestcolor}0.130&2$^\dag$&-\\
\midrule
Splatfacto-W  & \cellcolor{thirdbestc}26.87 & \cellcolor{bestcolor}0.932 & \cellcolor{thirdbestc}0.124 & 22.66 & 0.769 & 0.224 &22.53 & \cellcolor{secondbestc}0.876 & 0.158 & 1.05 & \textbf{40.2} \\
Splatfacto-W-A & \cellcolor{secondbestc}27.50 & \cellcolor{secondbestc}0.930 & 0.130 & 22.81 & 0.770 & 0.225 & \cellcolor{thirdbestc}22.62 & \cellcolor{secondbestc}0.876 & 0.156 & 0.83 & \textbf{58.8} \\
Splatfacto-W-T & 26.16 & 0.925 & 0.131 & 22.88 & \cellcolor{thirdbestc}0.772 & 0.228 & \cellcolor{secondbestc}22.78 & \cellcolor{bestcolor}0.878 & \cellcolor{thirdbestc}0.155 & 0.85 & \textbf{59.1} \\
\bottomrule
\end{tabular}
}
\caption{\textbf{Results on Three NeRF-W datasets.} 
We color each column as \textbf{\textcolor{bestcolor}{best}}, \textbf{\textcolor{secondbestc}{second best}}, and \textbf{\textcolor{thirdbestc}{third best}}.
* Results computed using a high-tier GPU~\cite{dahmani2024swag}.
$^\dag$ Results computed using a RTX3090~\cite{zhang2024gsw}.
Our results computed with a single RTX2080Ti. FPS are calculated without any caching. 
}
\label{table:results}
\end{table*}
\subsection{Implementation Details}
We minimize the \( L_1 \) loss combined with D-SSIM term and the alpha loss term to optimize the 3D Gaussians parameters alongside with the MLP \( F_\theta \) weights, appearance embedding and gaussian appearance features altogether. We train 65000 iterations on a single RTX2080Ti.

The appearance embedding is configured with 48 dimensions, while the gaussian appearance features are set to 72 dimensions. The architecture for the Appearance Model incorporates a three-layer MLP with a width of 256, and the Background Model employs a three-layer MLP with a width of 128.
\subsection{Quantitative Results}
We provide quantitative results using common rendering metrics: Peak Signal-to-Noise Ratio (PSNR), Structural Similarity Index Measure (SSIM), and Learned Perceptual Image Patch Similarity (LPIPS). Following the NeRF-W~\cite{martin2021nerf} evaluation approach, where only the embeddings for images are optimized during training, we optimize an embedding on the left half of each test image and report the metrics on the right half.

We train on all the images in the datasets and pick the same test image sets as NeRF-W~\cite{martin2021nerf} for evaluation. The final quantitative evaluation is provided in Table \ref{table:results}.

In this section, we also compare two variants of our method to analyze the contribution of each component of Splatfacto-W:

\begin{itemize}
    \item \textbf{Splatfacto-W-A}, a variation with only appearance model enabled.
    \item \textbf{Splatfacto-W-T}, a variation with only appearance model and robust mask enabled.
\end{itemize}

Our experiments demonstrate that our method yields competitive results. Remarkably, even without caching the SH coefficients for background and gaussian points, our method achieves real-time rendering at over 40 frames per second (fps) and supports dynamic appearance changes. With the current hyperparameters, our training process requires less than 6 GB of GPU memory and achieves the fastest performance on a single RTX 2080Ti, making training feasible on home computers. Additional image evaluation results are presented in Figure ~\ref{fig:scenes}.

\subsection{Background Modeling}
Our background model is also applicable in Splatfacto. Our method eliminates the majority of background floaters, providing greater background and depth consistency across different viewpoints without 2D guidance, as shown in Figure \ref{fig:bg}. More video results are available on our \href{https://kevinxu02.github.io/gsw.github.io/}{webpage}.

\section{Discussion}
Due to the lack of compression and understanding of image information by 2D models like U-Net, our method converges slowly on images with special lighting conditions, such as shadows and highlights caused by sunlight at specific times. Introducing additional networks and gaussian point features to learn the residuals between the image highlights and the current prediction results can alleviate this problem. However, this approach also introduces additional computational and storage overhead, which contradicts our initial objectives. Therefore, we ultimately did not adopt this method.

Although our masking strategy is effective in most cases and has minimal impact on training duration, the shadows and highlights in the aforementioned scenarios can result in significant loss, leading our model to overlook these parts and further complicating their convergence.

Another issue is that the our SH background model can only modeling low-frequency backgrounds, making it less effective at representing cloud portions, which also leads to the decline in PSNR.
\section{Conclusion}
In this paper, we introduced Splatfacto-W, a approach that significantly enhances the capabilities of 3D Gaussian Splatting (3DGS) for novel view synthesis in in-the-wild scenarios. By integrating latent appearance modeling, an efficient transient object handling mechanism, and a robust neural background model, our method addresses the limitations of existing approaches such as SWAG and GS-W.

Our experiments demonstrate that Splatfacto-W achieves better performance in terms of PSNR, SSIM, and LPIPS metrics across multiple challenging datasets, while also ensuring real-time rendering capabilities. The introduction of appearance features and robust masking strategies enables our model to effectively handle photometric variations and transient occluders, providing more consistent and high-quality scene reconstructions. Additionally, the neural background model ensures improved multiview consistency by accurately representing sky and background elements, eliminating the issues associated with background floaters and incorrect depth placements.

Despite these advancements, there remain challenges such as slow convergence in special lighting conditions and limitations in representing high-frequency background details. Future work will focus on addressing these issues by exploring more sophisticated neural architectures and additional network components to refine transient phenomena and enhance background modeling further.

\section*{Acknowledgments}
This work would not have been possible without the incredible support from the Nerfstudio team. Thanks to Professor Angjoo Kanazawa for her insightful guidance and mentorship. Special thanks to Justin Kerr for his pivotal role in hinting at this research direction, providing critical feedback on my ideas, and offering continuous guidance throughout the entire project. Thanks Ruilong Li for testing and optimizing appearance model for general datasets. And thanks to ShanghaiTech for offering the computing resources for running the experiments.

This project was funded in part by NSF:CNS-2235013 and IARPA DOI/IBC No. 140D0423C0035. JK is supported by NSF Fellowship.

{
    \small
    \bibliographystyle{ieeenat_fullname}
    \bibliography{main}
}
\end{multicols}


\end{document}